\documentclass{article}

\usepackage{arxiv}

\usepackage[utf8]{inputenc} 
\usepackage[T1]{fontenc}    
\usepackage{hyperref}       
\usepackage{url}            
\usepackage{booktabs}       
\usepackage{amsfonts}       
\usepackage{nicefrac}       
\usepackage{microtype}      
\usepackage{lipsum}
\usepackage{graphicx}
\usepackage{subfig}

\usepackage{authblk}

\usepackage[toc,page]{appendix}

\graphicspath{ {./images/} }

\newenvironment{boxed}
    {\begin{center}
    \begin{tabular}{|p{0.9\textwidth}|}
    \hline\\
    }
    {
    \\\\\hline
    \end{tabular} 
    \end{center}
    }

\title{MK-SQuIT: Synthesizing Questions using Iterative Template-filling}

\author[1\thanks{Completed during an internship at MeetKai Inc.}]{Benjamin A. Spiegel}
\author[2]{Vincent Cheong}
\author[2\thanks{Equal contribution.}]{James E. Kaplan}
\author[2$^\dagger$]{Anthony Sanchez}

\affil[1]{Brown University, Providence, RI 02912}
\affil[2]{MeetKai Inc., Marina Del Rey, CA 90292}
\affil[ ]{benjamin\_spiegel@brown.edu, \{vincent.cheong@, james@, anthony@\}meetkai.com}

\begin{document}
\maketitle
\begin{abstract}
The aim of this work is to create a framework for synthetically generating question/query pairs with as little human input as possible. These datasets can be used to train machine translation systems to convert natural language questions into queries, a useful tool that could allow for more natural access to database information. Existing methods of dataset generation require human input that scales linearly with the size of the dataset, resulting in small datasets. Aside from a short initial configuration task, no human input is required during the query generation process of our system. We leverage WikiData, a knowledge base of RDF triples, as a source for generating the main content of questions and queries. Using multiple layers of question templating we are able to sidestep some of the most challenging parts of query generation that have been handled by humans in previous methods; humans never have to modify, aggregate, inspect, annotate, or generate any questions or queries at any step in the process. Our system is easily configurable to multiple domains and can be modified to generate queries in natural languages other than English. We also present an example dataset of 110,000 question/query pairs across four WikiData domains. We then present a baseline model that we train using the dataset which shows promise in a commercial QA setting.

\end{abstract}


\section{Introduction}
Since the advent of BERT \cite{devlin2018bert} in 2018, substantial substantial research has been conducted into re-evaluating approaches to the question-answering task. Question Answering (or QA) systems can be generally considered as resolving a “context” and a “question” to an output “answer”. Where these systems often differ is in how they define these inputs and outputs. In open-domain systems, the context is a body of a text and the answer is a selection within the text that answers the question \cite{chen2017reading}. The context in a QA system can also be in the form of a table of data rather than documents of text \cite{yin20acl}. In generative QA systems, the same context is given but the model is tasked with generating an output response independently of a selection of text in the input \cite{yin2016neural}. In industry, a substantial amount of work has gone into generating “SQL” output rather than text to allow for querying from a database \cite{zhongSeq2SQL2017}. By generating a database query as the output, the model is capable of querying over much more data than could be provided in the form of a “context” text. 

In this work we focus on a generative approach similar to text2sql tasks, with the notable exception of generating SPARQL \cite{sparqlw3c} instead of SQL. By generating SPARQL a model is able to query over a “knowledge graph” instead of a traditional relational database. Numerous commercial services such as Bing, Google, and WolframAlpha utilize knowledge graphs to facilitate answers to user queries \cite{website:googlekb} \cite{noy2019industry}. Furthermore, open knowledge graphs, such as WikiData allow for both researchers and smaller entities to make use of a database consisting of over 1 billion facts \cite{vrandevcic2014wikidata}. This task is often called “KGQA” or Question Answering over Knowledge Graphs \cite{fu2020survey}. Numerous datasets have been created to facilitate this task, such as SimpleQuestions \cite{bordes2015large}, WebQuestions \cite{berant2013semantic}, QALD-9 \cite{ngomo20189th}, CQ2SPARQLOWL \cite{wisniewski2018competency} and most recently LC-QuAD 2.0 \cite{dubey2017lc2}. These datasets are often constructed utilizing a human in the loop at some stage. This can be in the form of annotation of asked queries, which is both expensive and non-standard, or in the creation of the questions through the use of crowdsourcing. While this approach can generate high quality datasets with proper controls, they suffer from the fact that they cannot be updated trivially with a quickly evolving knowledge graph of facts.  

This work differs notably from previous approaches to text2sparql datasets in that it takes a fully generative approach to the creation of the dataset. We find that even with an entirely automated dataset construction, we are able to achieve high user satisfaction on a commercial QA system trained on the generated dataset. In this work, we provide a modular framework for the construction of text2sparql datasets, a sample dataset to serve as a standard starting point, and finally a baseline model to demonstrate utilizing SOTA techniques to perform the KGQA task itself.

The key contributions\footnote{The code and sample dataset are available at \href{https://github.com/MeetKai/MK-SQuIT}{\texttt{https://github.com/MeetKai/MK-SQuIT}}} of this work are:
\begin{enumerate}
    \item Tooling
    \begin{enumerate}
        \item Dataset generation framework
    \end{enumerate}
    \item Dataset
    \begin{enumerate}
        \item 100k training set
        \item 5k easy test set
        \item 5k hard test set
    \end{enumerate}
    \item Model
    \begin{enumerate}
        \item Baseline BART model
    \end{enumerate}
\end{enumerate}

\section{Motivation}

Our work is most similar to LC-QuAD 2.0, which presented a dataset that leveraged crowdsourcing via Amazon Mechanical Turk to create a dataset of English questions paired with SPARQL queries. The authors of LC-QuAD 2.0 consider ten distinct types of questions:
\begin{enumerate}
    \item Single Fact: Who is the author of Harry Potter?
    \item Single Fact with Type: Kelly Clarkson was the winner of which show?
    \item Multi Fact: Who are the presidents of the US, who are from California?
    \item Fact with Qualifiers: What is the venue of Justin Bieber's marriage?
    \item Two Intention: Lionel Messi is a member of what sports team and how many matches has he played?
    \item Boolean: Is Peter Jackson the director of the Lord of the Rings?
    \item Count: How many episodes are in the television series Scrubs?
    \item Ranking: What is the country which has the highest population?
    \item String Operation: Give me all K-pop bands that start with the letter T.
    \item Temporal Aspect: How long did the Roman Empire last?
\end{enumerate}
Our work condenses their question types into a succinct list that is more amenable to our approach towards automated dataset generation:
\begin{enumerate}
    \item Single-entity (Single Fact, Single Fact with Type, Multi Fact, Fact with Qualifiers)
    \item Multi-entity (Boolean)
    \item Count
    \item Two Intention
    \item Ranking
    \item Filtering (String Operation, Temporal Aspect)
\end{enumerate}
In this work, we present methods for generating the first three question types, leaving the latter three for future work.  We define an entity as any primitive used in subject or object position of an RDF triple in the knowledge base. “Single-entity” questions contain a single entity and ask for some other related entity. “Multi-entity” questions contain two entities and ask if some specific relationship between them exists. “Count” questions contain a single entity and ask for the number of entities that satisfy a specific relationship to that entity. Each question type corresponds to a different basic SPARQL query template.
\begin{figure}[h]
  \centering
  \def\arraystretch{1.3}
  \begin{tabular}{ l | p{5cm} | p{7cm} }
    \textbf{Question Type} & \textbf{Question} & \textbf{Query} \\ \hline
    \texttt{Single-entity} & Who is the mother of the director of Pulp Fiction? & \texttt{SELECT ?end WHERE \{ [ Pulp Fiction ] wdt:P5 / wdt:P25 ?end . \}} \\ \hline
    \texttt{Multi-entity} & Is John Steinbeck the author of Green Eggs and Ham? & \texttt{ASK \{ BIND ( [ John Steinbeck ] as ?end ) . [ Green Eggs and Ham ] wdt:P50 ?end . \}} \\ \hline
    \texttt{Count} & How many awards does the producer of Fast and Furious have? & \texttt{SELECT ( COUNT ( DISTINCT ?end ) as ?endcount ) WHERE \{ [ Fast and Furious ] wdt:P162 / wdt:P166 ?end . \}}
  \end{tabular}
  \caption{Example questions and queries for "Single-entity," "Multi-entity," and "Count" question types.}
  \label{fig:fig1}
\end{figure}

\section{Dataset Generation Pipeline}

\begin{figure}[h]
  \centering
  \includegraphics[scale=0.1]{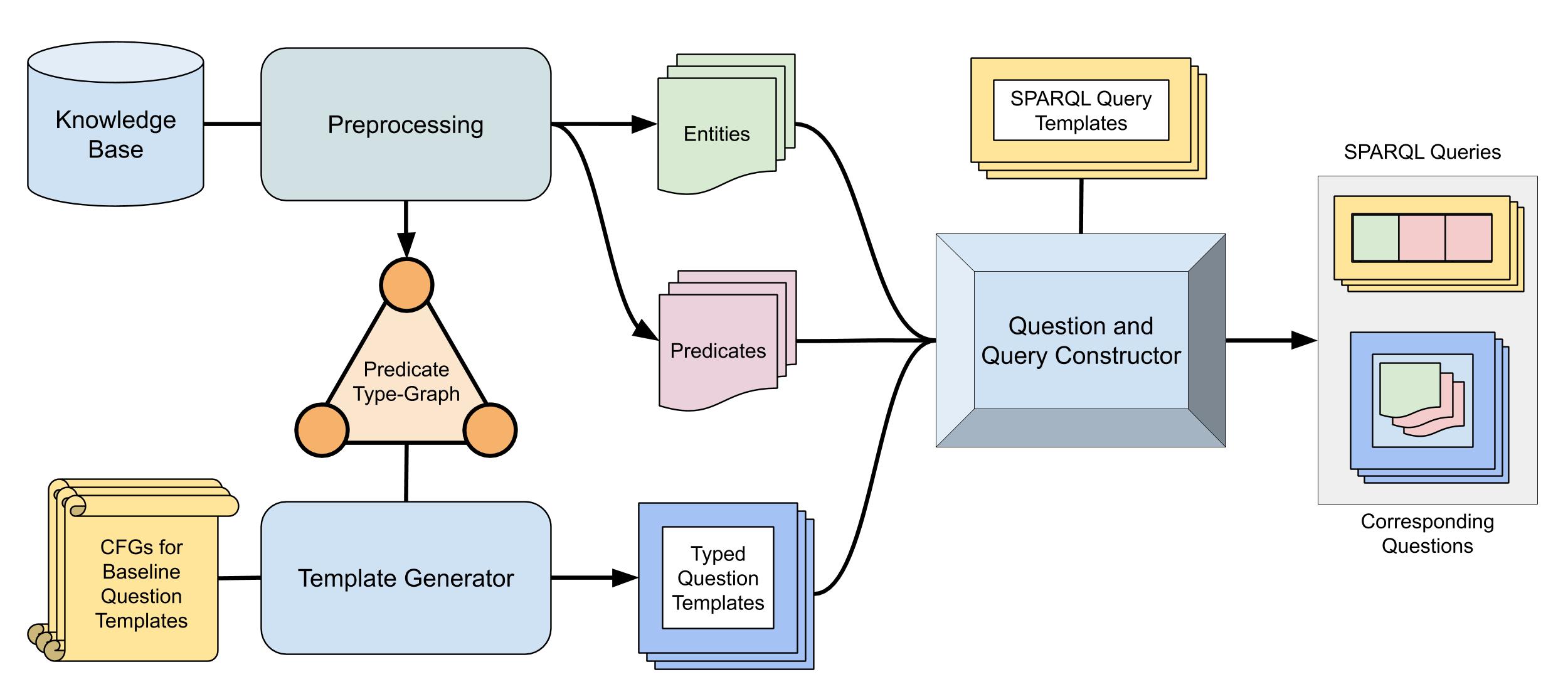}
  \caption{Pipeline Overview}
  \label{fig:fig2}
\end{figure}

\subsection{Designing the Question and Query Templates}

Context-free grammars, or CFGs, have had widespread use in classical linguistics and computer science for the formal definition of context-free languages \cite{chomsky1959algebraic}. In this work we utilize them to formally define each question type that we wish to cover in the dataset. The utilization of a CFG for question templates allows us to make use of their intrinsically recursive nature to generate deeply nested queries. The grammar productions of each CFG form the “baseline” templates for each question type. We have written three CFGs---one per question type---and generate productions that average a depth of 5, amounting to 51 total baseline question templates.

\begin{figure}[h]
  \centering
  \begin{boxed}
\textbf{Single-entity CFG} \\\\
S -> "[WH]" IS Q "?" \\
IS -> "is" | "was" \\
Q -> NOUN | VERB-ADP \\
NOUN -> "[THING]" | NOUN "'s [NOUN]" | "the [NOUN] of" NOUN \\
VERB-ADP -> NOUN "[VERB-ADP]"
\\\\\\
\textbf{Example Single-entity baseline template} \\\\
"[WH] was the [NOUN] of [THING] 's [NOUN] ?"
  \end{boxed}
  \caption{The CFG for Single-entity questions and an example baseline template produced from the CFG.}
  \label{fig:fig3}
\end{figure}

\subsection{Template Generator and Filler}

\begin{figure}[h]
  \centering
  \includegraphics[scale=0.1]{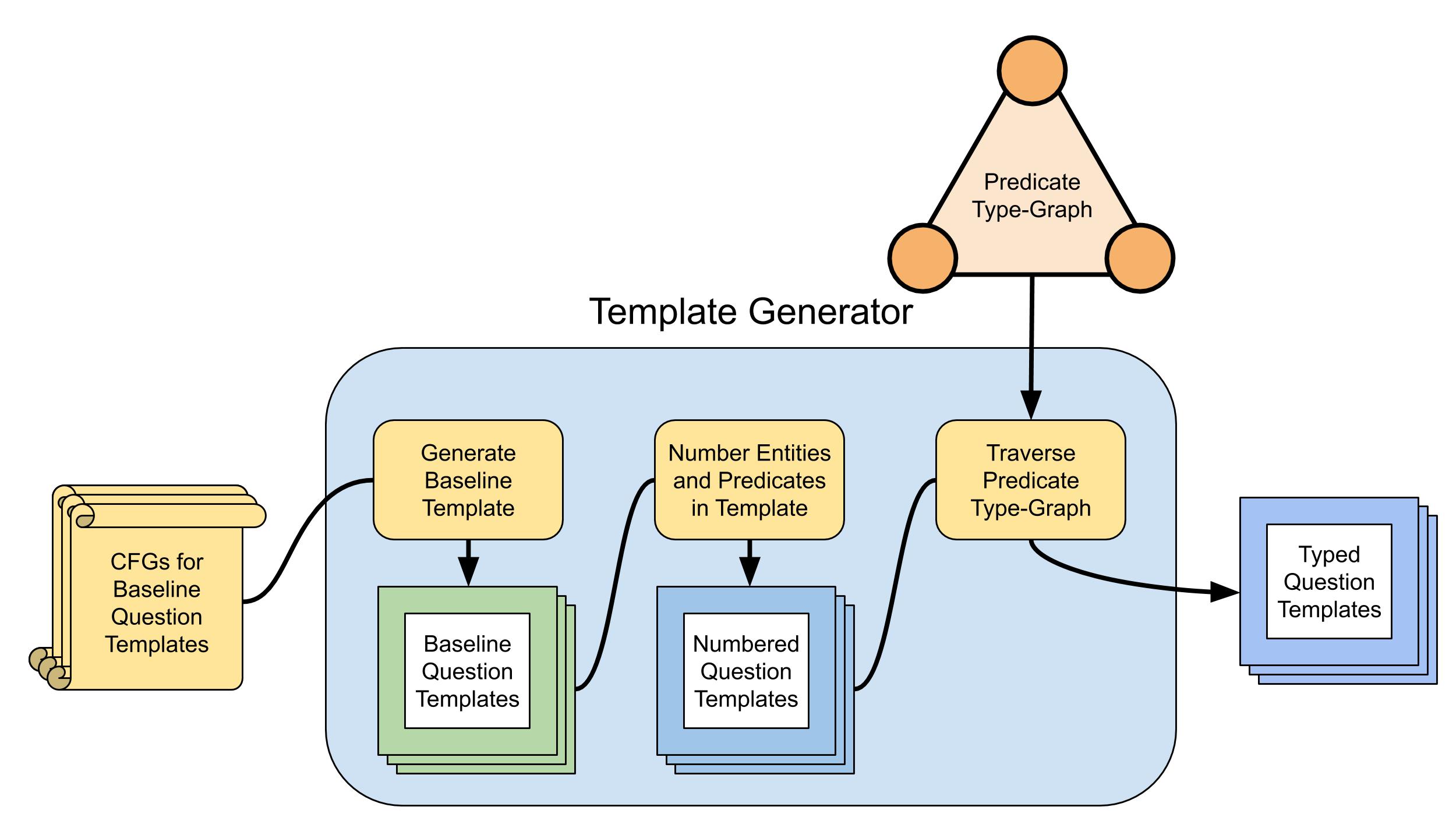}
  \caption{The Template Generator is responsible for generating semantically-aware templates that have predicates numbered according to the order they would appear in a SPARQL query.}
  \label{fig:fig4}
\end{figure}

The baseline templates are fed into a Template Generator module which adds additional constraints to templates to facilitate easy SPARQL generation and to control for possible semantic infelicity. The Template Generator first numbers the predicates in the baseline templates with the nesting order that they would take in their logical form \cite{liang2013lambda}. This is the same ordering that the predicates need to appear in the corresponding SPARQL query.

\begin{figure}[h]%
    \centering
    \subfloat[An example predicate type graph. For the Single-entity and Count queries, unidirectional traversals yield paths whose labeled edges are the type of a compatible predicate. For the Multi-Entity query types, a slightly more sophisticated bidirectional traversal is performed.]{{\includegraphics[scale=0.2]{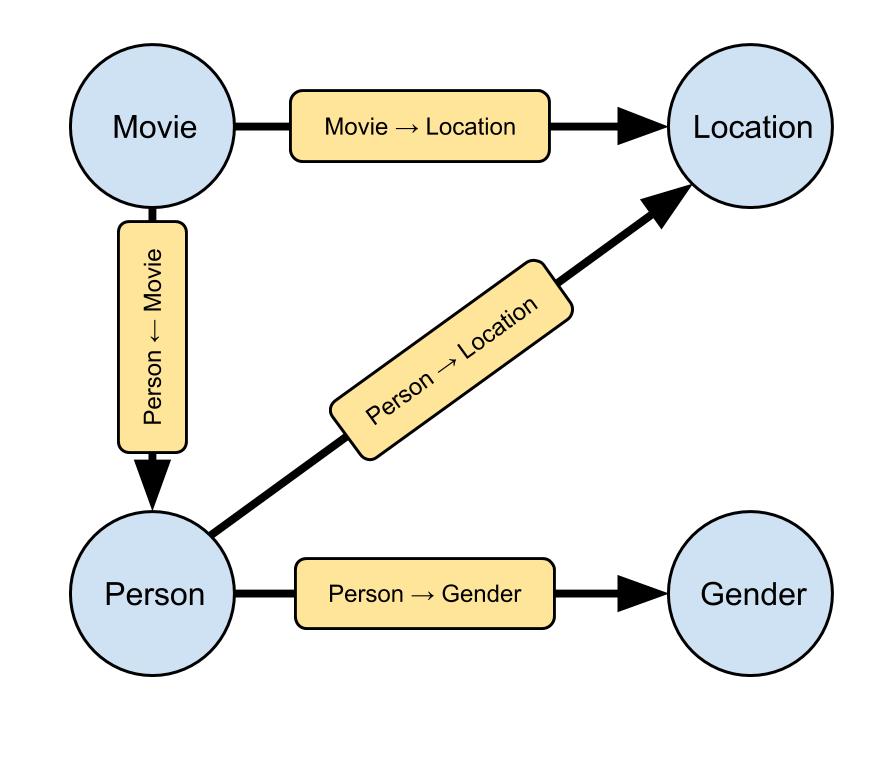} }}%
    \qquad
    \subfloat[The predicate ordering in the SPARQL query matches the order of the predicates in the typed template.]{{\includegraphics[scale=0.24]{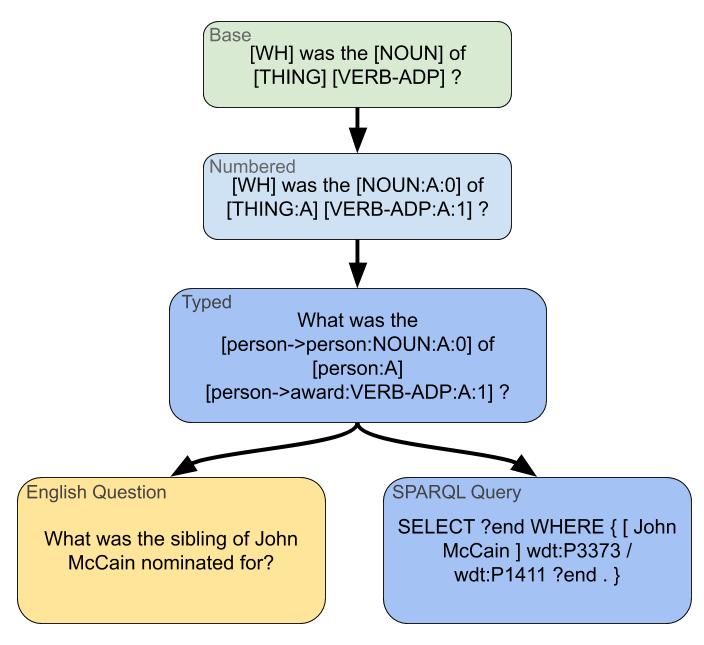} }}%
    \label{fig:fig5}
\end{figure}

Motivated by concepts in first-order logic, we treat each predicate as a function from one ontological type to another (e.g. “narrative location” is a function from television series to location). To enforce semantic felicity, it is necessary to ensure that the entities and predicates in the question have semantic types that correspond correctly. These semantic types are informed by the ontology type system used by the knowledge base. This step prevents questions like: "What is American Football's weight?" from being generated. \footnote{During the Preprocessing phase we take the responsibility of labeling each predicate with its ontological function type. In some knowledge bases this step might be done already.} Each chain of predicates in a question must satisfy the following conditions in order to be semantically felicitous:

\begin{enumerate}
    \item Each predicate in the predicate chain must take as an argument the type of entity that is output by the previous predicate.
    \item The first predicate must take as an argument the type of the entity in the main entity slot of the template.
\end{enumerate}

The construction of these chains is carried out through the use of a predicate graph, a directed graph of the legal predicate types in our Knowledge Base. While this graph was created based on WikiData, they could be trivially extended to any knowledge base. The predicate graph is best understood in the context of the “single-entity” templates. We begin the path in the graph at the randomly selected entity type node and make random traversals until the path reaches a length specified by the template. The predicates in the template are then labeled with the ontological types stored in the edges of the path. These labeled predicates can easily be independently sampled downstream during the Question and Query Construction phase, without any risk of compromising the semantic felicity of the generated questions.

\section{Sample Dataset: MKQA-1}

We extracted 133 distinct properties across four WikiData types: person, film, literary work, and television series. Properties that end with "\texttt{ ID}" were filtered out due to their abundance, as well as entities that were missing a label. Each property had an average of 8.46 alias labels, amounting to 1,019 property labels. For each WikiData type, we extracted 5,000 entities; disregarding duplicates, there were a total of 17,452 unique entities. Each entity had an average of 1.99 labels, amounting to a total of 34,074 unique entity labels. This data was used to generate 100k training examples.

\begin{figure}[h]%
    \centering
    \subfloat[The number of predicates extracted for each supported WikiData type. Some predicates are shared by multiple subject types. In total there are 133 unique predicates.]{{\includegraphics[width=7cm]{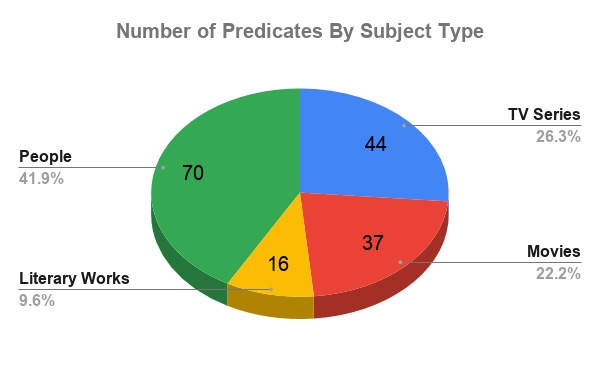} }}%
    \qquad
    \subfloat[A breakdown of the number of predicates by simplified POS tag. We elect to use NOUN and VERB-ADP in our example dataset, as our templates already cover the NOUN-ADP category by appending “of” to a NOUN category.]{{\includegraphics[width=7cm]{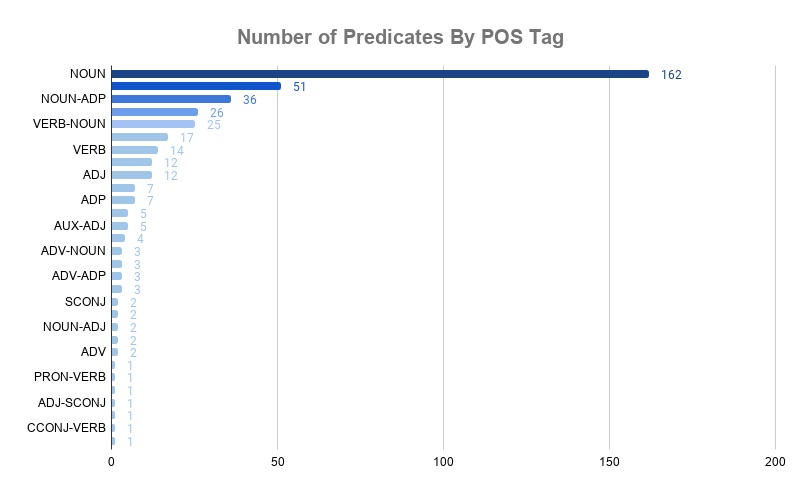} }}%
    \label{fig:fig6}
    \caption{Some statistics of the dataset}
\end{figure}

For the \textsc{test-easy} dataset, the same domain predicates were used in the generation process, but entities from a novel "Chemical" domain in WikiData were added and the baseline templates were longer and more complex. The \textsc{test-hard} dataset uses the same domain predicates as the training and easy test set, but uses entities exclusively from the chemical domain as well as even more complex baseline templates that contain additional filler words. The generated examples are then further processed with standard, open source augmentation tools \cite{ma2019nlpaug} for additional fuzzing. A total of 5k examples are generated for each test dataset.

\begin{figure}[h]
  \centering
  \includegraphics[scale=0.25]{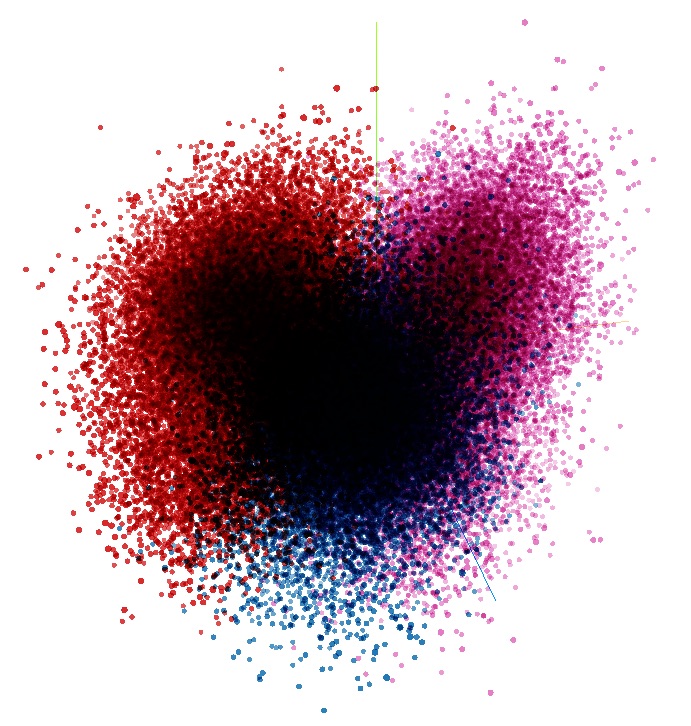}
  \caption{Visualized Embeddings using Tensorflow Projector. Each color represents a question type: blue for Single-entity, red for Multi-Entity, and pink for Count.}
  \label{fig:fig7}
\end{figure}

To explore the textual similarity of the dataset, we provide a visualization of the dataset through Tensorflow Projector (\url{https://projector.tensorflow.org/}) with sentence embeddings generated by a Universal Sentence Encoder \cite{cer-etal-2018-universal}.

\section{Baseline Model}

To evaluate the applicability of the dataset for a commercial QA system, we provide a simple BART \cite{lewis2019bart} model trained on our synthetic dataset and fine-tuned using NVIDIA's NeMo toolkit \cite{kuchaiev2019nemo}. We model the task as machine translation from natural language to SPARQL and use BART to initialize the weights of the network due to its pretrained ability to de-noise inputs. We also experimented with an Encoder-Decoder with a pretrained autoencoding and autoregressive model, but found that generation inference times were far slower, especially when query sequences were lengthy.

The BART model converges within 5 epochs with a learning rate of 4e-5. Sequence generation is performed with a greedy search, although a well configured beam search would be likely to improve performance. Generated output is then passed through minor post-processing to clean up query spacing. We have not adjusted any other hyperparameters or evaluated different model architectures, and instead present these findings as validation for the viability of utilizing a synthetic dataset for the KGQA task.

We find BLEU \cite{papineni2002bleu} and ROUGE \cite{lin2004rouge} to be good indicators of the model's performance. BART performs nearly flawlessly on the easy test set. For the hard test set we observe an expected decrease in scores reflecting that the model is challenged with more complex linguistic arrangements, noisy perturbations, and entities from unseen domains.

\begin{center}
  \def\arraystretch{1.3}
  \begin{tabular}{ | l | l | l | l | l | l | }
    \hline
    Dataset & BLEU & ROUGE-1 & ROUGE-2 & ROUGE-L & ROUGE-W \\ \hline
    \textsc{test-easy} & 0.98841 & 0.99581 & 0.99167 & 0.99581 & 0.71521 \\ \hline
    \textsc{test-hard} & 0.59669 & 0.78746 & 0.73099 & 0.78164 & 0.48497 \\
    \hline
  \end{tabular}
\end{center}

\section{Future Work}

\subsection*{Expansion into New Question Types and Natural Languages}
Expansion to new question types would require writing additional baseline template CFGs, predicate numbering functions, and perhaps making slight modifications to the Question and Query Constructor. Expansion to different natural languages will vary depending on the language, with some requiring only the modification of the baseline template CFGs, and others requiring large rewrites of the numbering functions. The quality of the queries is expected to decrease the more context-dependent the language is, since our generation framework assumes as little context-dependence as possible.

\subsection*{SPARQL Query Explainability}
To supplement this work, we used our synthetic dataset to train a model for translating from English to SPARQL. Alternatively, a model could be trained to do the reverse: translating a SPARQL query into an English question. This could be helpful for quickly deciphering SPARQL queries into English or other languages for the sake of readability.

\subsection*{Enhanced Coverage}
In the Motivation section, we mention that this work covers three of six possible question types. Writing the numbering and typing functions for the remaining three types is of interest to us for future work. One notable omission from our categorization of question types is one that queries triples where properties can also exist in subject or object position of an RDF triple. In WikiData, these higher-order properties are called qualifiers, and they are accessible via additional syntax.

\subsection*{Voice-Aware Augmentation}
While this dataset serves well for the task of translating English to SPARQL, it is not specifically tailored to process text that was produced from from a voice-to-text model. Voice-to-text models can bring a host of syntactic and semantic errors that can hamper the performance of downstream models in the pipeline. Recent work in Telephonetic Augmentation suggests that augmenting the dataset with these types of errors can yield substantial improvement when the model is fed voice input \cite{larson2019telephonetic}. We plan to augment our framework with a telephonetic fuzz module in future work.

\subsection*{Availability}
In order to encourage further work in the field, we open-source this framework for synthetic dataset generation in addition to the training and evaluation of our BART model under the MIT license. These can be found on GitHub at \url{https://github.com/MeetKai/MK-SQuIT}. A demo of the generation pipeline, baseline model, and dataset explorer will be available as a Docker container at NVIDIA NGC \url{https://ngc.nvidia.com/}.

\section{Conclusion}
In this work we presented a modular framework for the automated creation of synthetic English/SPARQL datasets by generating and refining question/query templates. We successfully evaluated the usefulness of this dataset by training a simple baseline model for an English-to-SPARQL machine translation task. We hope this work can serve as validation that QA systems can be trained without the need for crowdsourcing query/question data. Substantial future work is planned to aid in the development of SOTA KGQA systems in both research and commercial settings.

\bibliographystyle{unsrt}  
\bibliography{template}

\clearpage
\begin{appendices}
\section{Generated Dataset Samples}

\begin{figure}[h]
  \centering
  \def\arraystretch{1.3}
  \begin{tabular}{ l | p{5cm} | p{7cm} }
    \textbf{Question Type} & \textbf{Question} & \textbf{Query} \\ \hline
    \texttt{Single-entity} & What is Gisla saga's adaptation's place of origin? & \texttt{SELECT ?end WHERE \{ [ Gisla saga ] wdt:P4969 / wdt:P495 ?end . \}} \\
    & What is The Mummy playing in? & \texttt{SELECT ?end WHERE \{ [ The Mummy ] wdt:P840 ?end . \}} \\
    & What was the native language of Bite the Bullet's editor? & \texttt{SELECT ?end WHERE \{ [ Bite the Bullet ] wdt:P1040 / wdt:P103 ?end . \}} \\ \hline
    \texttt{Multi-entity} & Is Nick Cave the author of L'Assommoir? & \texttt{ASK \{ BIND ( [ Nick Cave ] as ?end ) . [ L'Assommoir ] wdt:P50 ?end . \}} \\
    & Is Lion of the Desert's recording location the place of residence of Cold Case's songwriter? & \texttt{ASK \{ [ Lion of the Desert ] wdt:P915 ?end . [ Cold Case ] wdt:P86 / wdt:P551 ?end . \}} \\
    & Is Francis Gillot the progeny of Look at Me's record producer? & \texttt{ASK \{ BIND ( [ Francis Gillot ] as ?end ) . [ Look at Me ] wdt:P162 / wdt:P40 ?end . \}} \\ \hline
    \texttt{Count} & What was the number of spoke language of Piotr Veselovsky? & \texttt{SELECT ( COUNT ( DISTINCT ?end ) as ?endcount ) WHERE \{ [ Piotr Veselovsky ] wdt:P1412 ?end . \}} \\
    & How many genre does Dem Täter auf der Spur have? & \texttt{SELECT ( COUNT ( DISTINCT ?end ) as ?endcount ) WHERE \{ [ Dem Täter auf der Spur ] wdt:P136 ?end . \}} \\
    & What is the number of writing languages of the screenwriter of Record? & \texttt{SELECT ( COUNT ( DISTINCT ?end ) as ?endcount ) WHERE \{ [ Record ] wdt:P58 / wdt:P6886 ?end . \}}
  \end{tabular}
  \caption{Samples of \textsc{Train} data for all question types. The felicities of the queries are partially dependent on how rigorously the knowledge base is typed. Most synthetic queries resolve to negative responses as the specific data does not exist within WikiData.}
  \label{fig:fig8}
\end{figure}

\begin{figure}[h]
  \centering
  \def\arraystretch{1.3}
  \begin{tabular}{ l | p{5cm} | p{7cm} }
    \textbf{Question Type} & \textbf{Question} & \textbf{Query} \\ \hline
    \texttt{Single-entity} & What was the place of origin of Thiepane? & \texttt{SELECT ?end WHERE \{ [ Thiepane ] wdt:P495 ?end . \}} \\
    & What was 2,4-MCPA? & \texttt{SELECT ?end WHERE \{ BIND ( [ 2,4-MCPA ] as ?end ) . \}} \\ \hline
    \texttt{Multi-entity} & Was the date of first publication of zinc phosphide the birthyear of 3,3'-bipyridine's authors? & \texttt{ASK \{ [ zinc phosphide ] wdt:P577 ?end . [ 3,3'-bipyridine ] wdt:P50 / wdt:P569 ?end . \}} \\
    & Is Francis Gillot the progeny of Look at Me's record producer? & \texttt{ASK \{ BIND ( [ Francis Gillot ] as ?end ) . [ Look at Me ] wdt:P162 / wdt:P40 ?end . \}} \\ \hline
    \texttt{Count} & Who was the number of writer of Monoxido de nitrogeno's derivative work? & \texttt{SELECT ( COUNT ( DISTINCT ?end ) as ?endcount ) WHERE \{ [ Monoxido de nitrogeno ] wdt:P4969 / wdt:P50 ?end . \}} \\
    & How many language of the reference does the derivative work of 1,5-cyclooctadiene have? & \texttt{SELECT ( COUNT ( DISTINCT ?end ) as ?endcount ) WHERE \{ [ 1,5-cyclooctadiene ] wdt:P4969 / wdt:P407 ?end . \}}
  \end{tabular}
  \caption{Samples of \textsc{Test-Easy} data for all question types. The chemical domain is added to increase the difficulty of mapping over entity values. The model only knows properties learned from the training set, so questions with chemicals are often nonsensical. }
  \label{fig:fig9}
\end{figure}

\begin{figure}[h]
  \centering
  \def\arraystretch{1.3}
  \begin{tabular}{ l | p{5cm} | p{7cm} }
    \textbf{Question Type} & \textbf{Question} & \textbf{Query} \\ \hline
    \texttt{Single-entity} & Hey how many distributor does ethyl cellosolve have? & \texttt{SELECT ( COUNT ( DISTINCT ?end ) as ?endcount ) WHERE \{ [ ethyl cellosolve ] wdt:P750 ?end . \}} \\ \hline
    \texttt{Multi-entity} & Hey was the awards of the mother of -24,25-dihydroxycholecalciferol's film crew member the win of the mom of Norepinephrine's favorite player's sisters and brothers? & \texttt{ASK \{ [ -24,25-dihydroxycholecalciferol ] wdt:P3092 / wdt:P25 / wdt:P166 ?end . [ Norepinephrine ] wdt:P737 / wdt:P3373 / wdt:P25 / wdt:P166 ?end . \}} \\ \hline
    \texttt{Count} & Do you know How much is the number of height of the step mother of the songwriter of Tabun? & \texttt{SELECT ( COUNT ( DISTINCT ?end ) as ?endcount ) WHERE \{ [ Tabun ] wdt:P86 / wdt:P3448 / wdt:P2048 ?end . \}}
  \end{tabular}
  \caption{Samples of \textsc{Test-Hard} data for all question types. In additional to chemical entities, questions incorporate noisy ASR/text-to-speech elements such as conversational filler and irregular capitalization. Templates are also produced with increased depth/complexity. }
  \label{fig:fig10}
\end{figure}

\end{appendices}

\end{document}